# Dendritic Cells for SYN Scan Detection

Julie Greensmith and Uwe Aickelin
School of Computer Science, University of Nottingham,
Nottingham, UK, NG8 1BB.
{jqg, uxa}@cs.nott.ac.uk

## ABSTRACT

Artificial immune systems have previously been applied to the problem of intrusion detection. The aim of this research is to develop an intrusion detection system based on the function of Dendritic Cells (DCs). DCs are antigen presenting cells and key to the activation of the human immune system, behaviour which has been abstracted to form the Dendritic Cell Algorithm (DCA). In algorithmic terms, individual DCs perform multi-sensor data fusion, asynchronously correlating the fused data signals with a secondary data stream. Aggregate output of a population of cells is analysed and forms the basis of an anomaly detection system. In this paper the DCA is applied to the detection of outgoing port scans using TCP SYN packets. Results show that detection can be achieved with the DCA, yet some false positives can be encountered when simultaneously scanning and using other network services. Suggestions are made for using adaptive signals to alleviate this uncovered problem.

Categories and Subject Descriptors: I.2 Computing Methodologies: Artificial Intelligence

## 1. INTRODUCTION

The Dendritic Cell Algorithm (DCA) is a recent addition to artificial immune systems (AIS), a collection of algorithms inspired by the human immune system. The DCA is based on current thinking in immunology, regarding the role of 'danger signals'[10] as activators of the immune system. It is shown experimentally that in the human immune system, Dendritic Cells (DCs) process danger signals and other indicators of damage and instruct the adaptive immune system to respond appropriately. In this paper we present an approach to intrusion detection inspired by the observed behaviour of natural dendritic cells.

In nature, DCs are sensitive to changes in concentration of different signals derived from their tissue environment. DCs combine these signals internally to produce their own output signals in combination with location markers in the form of antigen[11]. The signal combination procedure is facilitated through a mechanism known as signal transduction. The signals received from the tissue during antigen collection determines the context in which the antigen is presented to the adaptive arm of the immune system. The outcome is either tolerance or activation towards entities expressing antigen of the same structure as the presented antigen. DCs form ideal inspiration for an artificial immune system based intrusion detection algorithm as they are a key cell in this biological decision.

The DCA is not the first AIS algorithm applied to intrusion detection. In fact, intrusion detection systems (IDS) were amongst the first applications of any AIS, based on the premise of combating computer viruses with a computer immune system. Negative selection [7] has been used with some success, but is plagued by problems surrounding scaling and false positives[14]. The premise of the Danger Project [1] is to alleviate the problems encountered with negative selection through the incorporation of danger theory based immunology to AIS. The DCA is one of the significant research outcomes of the Danger Project, alongside the libtissue framework[16] and developments in DC biology[17].

The aim of this paper is to expand on previous work using the Dendritic cell algorithm [5], by applying the algorithm to realistic port scan detection and observing the effects on the system. This can be used to gain insight into the behaviour of the DCA under different conditions and the development of a generalised signal selection schema.

This paper describes the application of the DCA to the detection of port scans based on the sending of SYN packets. This type of scan is termed a SYN scan. Section 2 contains background information regarding the use of AIS in computer security, an overview of the immunology used as inspiration, and an overview of the DCA itself. General rules for signal selection are outlined in Section 3, with examples of signals used for the purpose of scan detection. Experiments demonstrating the use of the DCA for SYN scan detection are outlined in Section 4. Results show the effects of different scanning scenarios on the detection capabilities of the algorithm. The final sections include a discussion of these results and their implications for the future of this algorithm.

## 2. BACKGROUND

In this section we present information regarding the use of AIS within Intrusion detection, a summary of Dendritic cell biology and the fundamentals of the Dendritic Cell Algorithm.

### 2.1 Computer Security and AIS

Intrusion detection in computer security is the detection of unauthorised use and abuse of computer systems and networks. The majority of techniques in IDS rely on signature-based misuse systems, where patterns of known malicious behaviour are stored in a database and are compared against observed patterns at run-time[13]. This approach can lead to false negative errors as the signature base must be constantly updated in order to provide adequate protection. Another approach tried by the IDS community is anomaly detection. In this paradigm, a profile of good or normal behaviour is created from training data. Deviations from normal result in the generation of alerts. Anomaly detection systems can be prone to false positive errors, detecting normal actions as anomalous, because normal is difficult to define and can change over time. Early AIS were developed for the purpose of detecting intruders in the context of computer security[3]. The method employed to achieve protection against breaches in computer security was the negative selection (NS) algorithm [7].

Extensive amounts of work have been performed with this algorithm, spanning over a decade of research within AIS[2]. In particular, the NS algorithm has been applied to the detection of anomalous connections between computers[7]. This anomaly detection style system used a supervised learning paradigm. Network connections are represented as bit-strings, and a profile of normal strings is created as training data. These positive examples of normal are shown to a set of detectors, who are assigned randomly generated strings. Each detector is matched against each training item for similarity assessment. Should a detector match a sufficient number of normal strings, it is deleted from the detector set. This filtering results in a set of detectors tuned to detect strings which fall outside this normal set. This functions in a similar manner to mechanisms shown in classical immunology based on the self-nonself paradigm.

The NS approach has a number of problems, highlighted by various researchers within AIS and proved both empirically and theoretically. Firstly the algorithm does not scale as well as expected[8]. This is due to the randomisation process associated with the generation of the detector set. As the size of the detector space increases, the number of detectors needed to cover the space increases exponentially. Additionally, a higher rate of false positives was shown than expected [14], despite attempts to improve the representation and the addition of features such as user interaction based 'co-stimulation' techniques. The false positive problem arises due to the initial static definition of normal. What is 'normal' changes over time, as new previously unseen connections are made and once trusted connections can become subverted for malicious purposes.

In 2003, Aickelin et al [1] outlined the Danger Project, describing the application of the 'danger theory' to intrusion detection systems. The authors suggested a system of detection based around the presence or absence of danger signals as opposed to the pattern-matching based approach used in negative selection. Danger signals released as a result of dying cells indicate damage, and stimulate the immune system. It was proposed that a system which could differentiate between data collected in a dangerous context with data collected in a safe context. It was suggested that some of the problems with false positives could be alleviated through the incorporation of these two contexts for the purpose of IDS. As dendritic cells are a key cell in the translation of danger signals, they have formed a central part in the development of the danger based IDS, described in this paper.

## 2.2 Dendritic Cells

In this section a brief overview of the biological principles used in the Dendritic Cell Algorithm are introduced. For more detailed discussion of DC biology, please refer to [9] or [4].

In the human body, DCs have a dual role, as garbage collectors for tissue debris and as commanders of the adaptive immune system. DCs belong to the innate immune system, and do not have the adaptive capability of the lymphocytes of the adaptive immune system. DCs exist in various states of differentiation, which determines their exact function. Modulations between the different states are dependent upon the receipt of signals while in the initial or immature state. The signals in question are derived from numerous sources, including pathogens, from healthy dying cells, from damaged cells and from inflammation. Each DC has the capability to combine the relative proportions of input signals to produce its own set of output signals. Input signals are categorised based on their origin:

PAMPs: Pathogenic associated molecular patterns are proteins expressed exclusively by bacteria, which can be detected by DCs and result in immune activation. The presence of PAMPS usually indicates an anomalous situation.

Danger signals: Signals produced as a result of unplanned necrotic cell death. On damage to a cell, the chaotic breakdown of internal components form danger signals which accumulate in tissue. DCs are sensitive to changes in danger signal concentration. The presence of danger signals may or may not indicate an anomalous situation, however the probablility of an anomaly is higher than under normal circumstances.

Safe signals: Signals produced via the process of normal cell death, namely apoptosis. Cells must apoptose for regulatory reasons, and the tightly controlled process results in the release of various signals into the tissue. These 'safe signals' result in immune suppression. The presence of safe signals almost certainly indicate that no anomalies are present.

Inflammation: Various immune-stimulating molecules can be released as a result of injury. Inflammatory signals and the process of inflammation is not enough to stimulate DCs alone, but can amplify the effects of the other three categories of signal. It is not possible to say whether an anomaly is more or less likely if inflammatory signals are present. However, their presence amplifies the above three signals.

Dendritic cells act as natural data fusion agents, producing various output signals in response to the receipt of differing combinations of input signal. The relative concentration of output signal is used to determine the exact state of differentiation, expressed by the production of two molecules, namely IL-12 and IL-10. In their immature state, dendritic cells collect antigen within the tissue compartment. During this phase they are exposed to varying concentrations of the input signals. Exposure to PAMPs, danger signals and safe signals causes the increased production of costimulatory molecules, and a resulting removal from the tissue and migration to a local lymph node.

DCs translate the signal information received in the tissue into a context for antigen presentation, i.e. is the antigen presented in an overal 'normal' or 'anomalous' context. The antigen collected while in the immature phase is expressed on the surface of the DC. Whilst in the lymph node, DCs seek out T-lymphocytes (T-cell) and attempt to bind expressed antigen with the T-cells variable region receptor. T-cells with a high enough affinity for the presented antigen are influenced by the output signals of the DC. DCs exposed to predominantly PAMPs and danger signals are termed 'mature DCs'; they produce mature DC output signals, IL-12, which activate the bound T-cells. This links the activation of T-cells to the potential suspect antigen present in tissue when intruders and damage are evident. Once activation has been achieved the T-cell travels back to the tissue to seek out any entity displaying a matching antigen. Conversely, if the DC is exposed to predominantly safe signals, antigens are presented in a safe context, as little damage is evident when the antigen is collected. This induced state of differentiation is termed semi-mature. In this state the DC produces IL-10 which has the ability to de-activate T-cells. . If the matching T-cell encounters an entity expressing this antigen, no response is mounted. The balance between the signals is translated via the signal processing and correlation ability of these cells. The overall immune system response is based on the systemic maturation state average of the whole DC population. An abstract view of this process is presented in Figure 1.

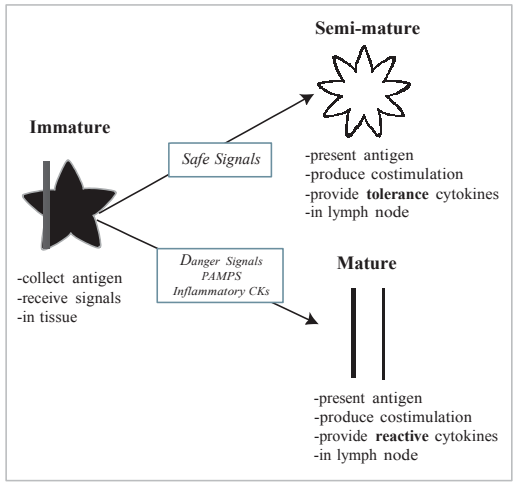

Figure 1: An abstract view of DC maturation and signals required for differentiation. CKs denote cytokines, molecular messengers between immune system cells.

### 2.3 The Dendritic Cell Algorithm

The Dendritic Cell Algorithm (DCA) was first introduced by Greensmith et al [4] in 2005. It has since been applied to two-class classification of a static machine learning dataset[4], the detection of small-scale port scans, under both off-line conditions[5] and in real-time experiments [6]. It represents a shift in focus within the field of AIS, from algorithms based solely on the adaptive immune system function to those incorporating metaphors derived from the innate immune system. This has paralleled similar trends in immunology, where for decades it was believed that the immune system used a pattern based system to identify pathogens. Opposition to this theory in the light of volumes of opposing evidence stimulated research of the innate driven mechanism demonstrated via DC behaviour and how this integrates with the concepts of classical immunology. In a similar manner, the DCA abandons the use of pattern matching to classify antigen, as previously used in the Negative Selection algorithm [7]. As a result it does not suffer the scaling problems outlined for negative selection. A brief description of the algorithm follows below, with a detailed description and its implementation given in Greensmith et al [5].

The DCA is a population based system, with each agent in the system represented as a cell. Each cell has the capacity to collect data items, termed antigen, and the processing of values of input signal. The combination of the input signals forms cumulative output signals of the DCs. The population of cells is used to correlate co-occurring and disparate data sources, effectively combining the 'suspect' data (antigen) with 'evidence' in the form of signals.

The algorithm uses the notion of tissue, which supports the initial processing of data, as implicated in Twycross and Aickelin [16]. Two 'compartments' are necessary, one for data collection and processing termed tissue, and one for the analysis of antigen termed a 'Lymph node'. A diagram of the DCA is presented in Figure 2.

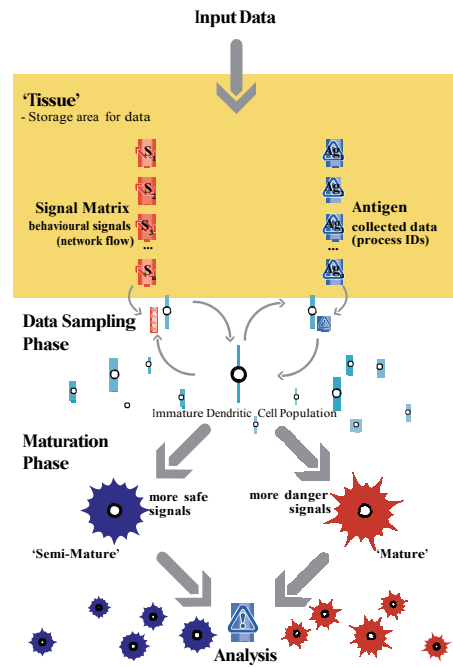

Figure 2: Illustration of the DCA showing data input, continuous sampling, the maturation process and antigen analysis.

In the tissue compartment, input data is stored for collection by the DC population. Antigens are stored in an antigen vector, with signals stored in a signal matrix. The population of DCs is stored as an array of objects. Multiple signals of different categories can be used as input and stored in the matrix. The signal matrix facilitates the representation of signals within the system, providing the interface between raw values of input data and signal values for the use of the DC population. Signals and antigen are streamed to the tissue, with both storage data structures updated upon the arrival of new data. Cells in the sampling population are updated once per second. During this update, each cell selects 10 indices within the antigen vector, and transfers any antigen contained within the vector to the cell's own antigen store. Once the antigen vector has been sampled, values from the signal matrix are copied to the cells internal signal store. Cumulative output signal values are updated each time the signal matrix is visited. A schematic representation of the signal processing equation is shown in Figure 3.

The output signal value representing the costimulatory molecules (CSMs) is used as a marker of maturation, enforcing a limit on the time a cell spends sampling before migrating to the lymph node. The value for CSM is incremented in proportion to the quantity of input signals received. The input signals are combined to form CSMs using a simple weighted sum. Weights for this equation are shown in previous work [5]. They have been derived from immunological observations [17] and were refined based on a sensitivity analysis performed in previous work [6]. Once the value of CSM is greater than the cells migration threshold, the cell is removed from the sampling population and is transferred to the Lymph node compartment.

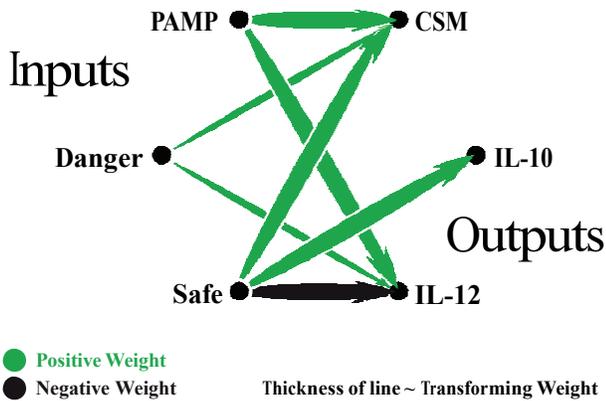

Figure 3: A schematic diagram of the signal processing equation used by every DC to fuse input signals and derive output signals.

Assessment of the output signals of DCs (either IL-10 or IL-12) in the lymph node is used to form the context of the collected antigen. The two remaining output cytokines are assessed. A high value of collected PAMPs and danger signals with a low value of safe signal is likely to result in an increase in mature output signal. Conversely a high value of safe signal will result in a high value of semi mature output signal. The values for semi mature and mature output signals are compared. The context of the DC is given by the output signal with the greatest value, with the assignment of context of 0 for semi-mature and 1 for mature. In the event of a tie, the cell is given a context of 0.

Antigens collected by the DC are printed to a log file, in combination with the context of the cell. An average context can be calculated for antigens of identical value or structure (type of antigen). The total fraction of mature antigen, per type of antigen, is derived forming the mature context antigen value or MCAV coefficient. The nearer a MCAV is to 1, the more likely the antigen is anomalous, as it was frequently collected in a context with high values of danger signals and PAMPs repeatedly. A confidence metric for the MCAV is derived using the total number of antigen presented per type of antigen. The higher the confidence value, the greater the probability of the MCAV being an accurate representation of the processed data. The uses of the antigen confidence indicator are shown in Section 5.

## 3. SYN PORT SCAN DETECTION

Previous use of the DCA involved the detection of a short and simple ping scan, based on the ICMP 'ping' protocol. To challenge the capability of the DCA, in this paper the algorithm is applied to the detection of the more complicated port scan, the TCP SYN scan. This is a commonly used type of scan, which leaves no trace in the normal system logs. Port scans form an ideal model of an intrusion, and techniques applied to scan detection can also be used to detect scanning worms. Early detection of scanning can prevent more serious attacks, as it is a tool crucial to the information discovery stage of intrusions. In this paper, we aim to detect a SYN scan launched from a victim machine, where the DCA is used to monitor the behaviour of the victim. This forms a scenario representing a scan performed by an insider, namely a legitimate user of the system who uses the system in an unauthorised manner.

The SYN scan itself is used to determine which ports are open and which services are running on specified hosts. Unlike the default TCP Connect scan, the SYN scan leaves no trace in normal system logs, as the TCP '3-way handshake' is left incomplete. SYN scans involve sending TCP packets to IP addresses specified at the command line of the scan program. The scanning machine sends a SYN packet to each address, and uses the information retrieved from the scanned remote machines to characterise the network. If a SYN packet is sent to a closed port, the remote machine responds by sending a TCP reset (RST) packet back to the scanning machine. Conversely, if the port is open, the remote machine responds with a TCP SYN-ACK packet. The scanning machine then terminates the potential TCP connection by sending a RST packet to the remote machine. As the 3 way handshake is not completed, no actual connection is made to the remote machine.

### 3.1 Signals

In vivo, DCs combine input signals in the form of concentration of molecules, translated through a network of receptors, signal transduction mechanisms and gene regulatory processes. Natural DCs are sensitive to changes in their environment, described as the chemical content of tissue which the DCs can sense through their expressed receptors. In a similar manner, the DCs used in the DCA are sensitive to changes of value within the signal matrix. This system relies on correct mapping of signals through examining the nature of the input data, and the assignment of correct weightings. To assist the signal selection process, a set of general rules for correct mapping are defined. Previous experiments have shown that modifications to the mappings of the different signals can lead to the generation of false positive errors [5], though robustness is shown if PAMPs and danger signals are incorrectly mapped. Input signals are abstracted from the general biological principles outlined in the previous section:

- PAMPs: A signature of abnormal behaviour, e.g. errors per second. An increase in this signal is associated with a high confidence of abnormality.

- Danger Signal: A measure of an attribute which increases in value to indicate an abnormality e.g. number of network packets per second. Low values of this signal may not be anomalous, giving a high value a moderate confidence of indicating abnormality.

- Safe Signal: A measure which increases value in conjunction with observed normal behaviour e.g. a low rate of change of packet sending. This is a confident indicator of normal, predictable or steady-state system behaviour. This signal is used to counteract the effects of PAMPs and danger signals.

- Inflammation: A general signal of system distress, which is insufficient to cause any maturation in the absence of other signals e.g. many users logged into a system remotely. Used to amplify the effects of the other signals.

For the detection of SYN scans, seven signals are derived from behavioural attributes of the monitored machine: two PAMPs, two danger signals, two safe signals and one inflammatory signal. Having two signals in each category should make the DCA more robust against random network fluctuations. As the inflammatory signals is observed locally, one signal should be sufficient. The PAMP signals are both taken from data sources which indicate a scan specifically. Danger signals are derived from attributes which represent changes in behaviour. Safe signals are also derived from changes in behaviour, but high safe signal values are shown when the changes are small in magnitude. The inflammatory signal is simplified as a binary signal i.e. inflammation present or not. All PAMPs, dangers and safe signals are normalised within a range of 0 and 100 to facilitate further processing.

PAMP-1 is the number of ICMP 'destination unreachable' error messages received per second. Scanning IP addresses which are not attached to a running machine or machines which are firewalled against ICMP packets generate these error messages. This signal was proved useful in detecting ping scans, and may also be useful in the detection of SYN scans, as an initial ping scan is performed to find running hosts. In this experiment, the number of ICMP messages generated was significantly less than observed with a ping scan. To account for this, normalisation of this signal includes multiplying the raw signal value by 5, capped at a value of 100.

PAMP-2 is the number of TCP reset packets sent and received per second. Due to the nature of the scan, a volume of RST packets are created in both port status cases; they are generated from the scanning machine if ports are open, and are generated by the remote machines if ports are closed. RST packets are not usually present in any considerable volume, so their increased frequency is a likely sign of scanning activity. This signal is normalised linearly, with a maximum cap set at 100 RSTs per second.

The first danger signal (DS-1) is derived from the number of network packets sent per second. Previous experiments with this signal [5] indicate it is useful for the detection of outbound scans. A different approach is taken for the normalisation of this signal. A sigmoid function is used to emphasise the differences in observed rate, making the range of 100 to 700 packets per second more sensitive. This function makes the system less sensitive to fluctuations under 100 packets per second, whilst keeping the sensitivity of the higher values. A cap is set at 1000 packets per second. The resulting signal range is between 0 and 100.

DS-2 is derived from the ratio of TCP packets to all other packets processed by the network card of the scanning machine. This signal is used as during SYN scans there is a burst of traffic comprised of almost entirely TCP type packets. The ratio is noramlised through multiplication by 100, to give this signal the same range as DS-1.

Safe signals are implemented to counteract the effects of the other signals, hopefully reducing the number of false positive antigens. The first safe signal (SS-1) is applied as described in [5] and encapsulates the rate of change of sending of network packets. High values of this signal are achieved if the rate of change is small and vice versa. This implies that a large volume of packets can be legitimate, as long as the rate at which the packets are sent remains constant.

The second safe signal (SS-2) is based on the observation that during SYN scans the average network packet size drops to a size of 40 bytes. Observations under normal conditions show that the average packet size is within a range of 70 and 90 bytes. A step function is implemented to derive this signal, with raw values between 40 and 45 bytes given a SS-2 value of 0, 46-50 bytes a value of 10, 51-60 bytes a value of 50, and over 61 bytes a value of 100. Preliminary experiments showed that a moving average is needed to increase the sensitivity of this signal. This average is created over a 60 second period.

The inflammatory signal is binary and is based on remote root log-ins. If a remote root log-in is detected this signal equals one, acting as a multiplier for the other signals.

### 3.2 Antigen

The signals have been selected for the detection of SYN scans, based on observed changes in machine behaviour during a scan. Hence, the signals chosen in this paper differ from those in our previous research. However, the antigen used for ping scan detection is also suitable for the detection of SYN scans. Process identification numbers (PIDs) generated each time a system call is made form the antigen. All remote sessions facilitated by ssh are monitored for this experiment. Using multiple system calls with identical PIDs allows for the aggregate antigen sampling method, described in Section 2. This allows for the detection of exactly which process was active when changes in signal values are observed. This technique is a form of process anomaly detection, but the actual structure of the PID is not important in terms of its classification, i.e. no pattern matching is performed, PIDs simply represent labels to identify processes.

## 4. EXPERIMENTS

The aim of this experiment is to apply the DCA to the detection of a SYN scan, launched from the machine which the algorithm is monitoring. Two datasets are used for this purpose: passive normal and active normal. The passive normal dataset emulates a 'night time' scan, while the machine is not being actively used. It should be relatively easy to detect anomalous behaviour in this data set. The active normal dataset includes simultaneous web-traffic and scanning processes.

The active normal dataset is 7000 seconds in duration, with 'normal' antigen generated by running Firefox over a remote SSH session. During browsing, multiple downloads, chat sessions and the receipt of e-mail occurred representing different patterns of network behaviour. The passive normal dataset comprises of a nmap scan and its pts-ssh demon parent. Both datasets contain processes which were invoked as a result of running a remote SSH session to run the scan, logged in using a root password. Both datasets contain a SYN scan of all ports using 254 IP addresses. Approximately 70 hosts were available at any time during the scan. As the scanned machines are part of a university network and the availability of the machines is beyond our control. The scan performed in both datasets is a stealth SYN scan, with a fast probe sending rate ( <0.1 sec per probe), facilitated through the use of the popular scanning tool, nmap[12]. The command used is "nmap -sS -v xxx.xxx.xxx.1-254. Once these initial datasets are created, a 'replay client' is used to process the same data repeatedly for different experiments, even though the system is designed to and does work in real-time.

Previous experience using the DCA has shown that there is little deviation in the output of the algorithm run on the same dataset, rendering repeats of identical experiments unnecessary. This is due to the sheer volume of input antigen sampled and the stochastic nature of the sampling process. System parameters for these experiments are as follows : number of signal categories = 4; number of signals per category = 2; tissue antigen storage = 500; number of cells = 100; number of antigen taken by DC in 1 update = 10; number of antigen stored by a DC = 50; and the number of DC output signals = 3. The MCAV coefficient is calculated for every 10000 antigen presented, a number derived during preliminary investigations. In terms of assessment, the PIDs with the highest volume of antigen output are used as the processes of interest. For the passive normal dataset these processes are the nmap scan process and the ssh demon. The processes of interest for the active normal dataset include the nmap scan, pts process and the Firefox browser and its children. Graphs are generated showing the MCAV for each process of interest per 10000 antigens presented, for the duration of the experiments. We expect higher values of MCAV for the nmap process and its parent process, the ssh demon, than for the Firefox browser or the bash shell.

All experiments are performed on an AMD Athlon 1GHz Debian linux machine (kernel 2.4.10). The algorithm is implemented within the `libtissue` framework[15], implemented in C (gcc 4.0.2) with interprocess communication facilitated by the SCTP protocol. All signals are derived using signal collection scripts, with values taken from the 'proc' filesystem (PAMP-1, DS-1, SS-1, I), the tcp-stat linux utility (D2, SS-2) and a custom developed packet sniffer (PAMP-2).

## 4.1 Results

Figures 4 and 5 show the input signals for the active normal (AN) and passive normal (PN) datasets respectively. In reality, both sets of signals are extremely noisy, and the figures depicted are smoothed representations of the actual signal values used in the experiments. Inflammation is a binary signal an is not represented on these figures. The AN signals are more variable than the PN signals, as many more processes run during the AN session. In the AN session, the duration of the nmap scan is approximately 6000 seconds, with the scan initiating at 651s. Signals PAMP-1, PAMP-2 and DS-2 clearly change for the duration of the scan. The remaining signals are less clear, though some evidence of changes throughout the scan duration is shown. The changes are transient and localised in particular to the beginning of the scan, when the majority of probes are sent to other hosts.

The signals of the PN dataset are less noisy. Analysis of input antigen confirms nearly 99% of these antigens belong to the anomalous pts and nmap processes. PAMP-1 and PAMP-2 are responsive to the scan, as shown by their rapid decline towards the end of the scanning period, at 5500s. Changes in DS1 are more pronounced in the PN dataset, yet the magnitude of this signal is smaller than expected. DS-2 appears to be highly correlated with the scan, yielding values of over 20 throughout the scan duration. SS-1 performs poorly, and only decreases in response to the scan in a few select places. SS-2 falls sharply in the middle of the scan, as predicted, but otherwise remains at a constant level of 60 even after the scan has finished.

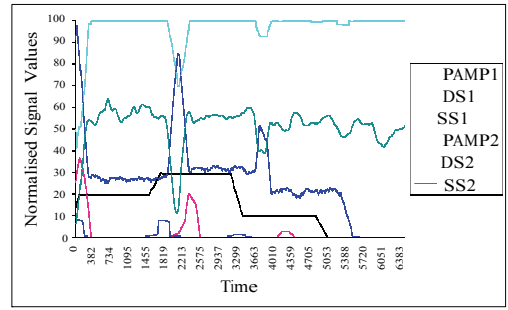

Figure 5: Simplified sketch of the input signals comprising the passive normal dataset. Inflammation is not represented.

every 10000 antigens. In Figures 6 and 7 each point represents the MCAV over 10000 antigens per process, for each process of interest. Figure 6 shows the MCAV output for PN. High values of over 0.5 are shown consistently for both nmap and pts in this experiment. This is evident for the first 40000 antigens. The MCAV values for the remainder of the scan are low, as at this point the scan slows to such extent that the behaviour of the machine remains constant, causing little change in signals and resulting in low MCAV values.

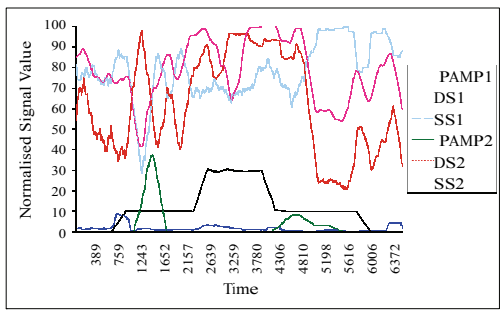

Figure 4: Simplified sketch of the input signals comprising the active normal dataset. Inflammation is not represented.

The antigen log file for the output of the DCA in both experiments (AN and PN) is partitioned and MCAVs recalculated after

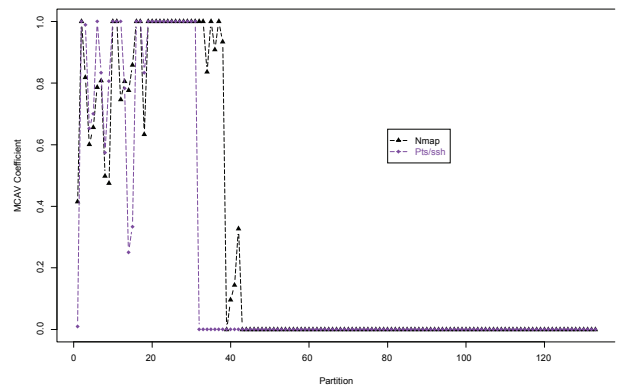

Figure 6: MCAV per 1000 antigens for the passive normal dataset. Nmap and Pts processes detected and represented.

Figure 7 presents the results of the active normal datasets. High MCAVs for the nmap and pts processes are evident throughout the scan duration, indicating successful detection. However, a number of firefox antigens also have high values of MCAV, reaching the maximum value of 1 as the scan is performed. Similarly, the nmap and firefox mean MCAV across the entire session is 0.16 for both processes, with identical standard deviations of 0.31. It appears that it is not possible to separate the two active processes if normal and attack processes run concurrently.

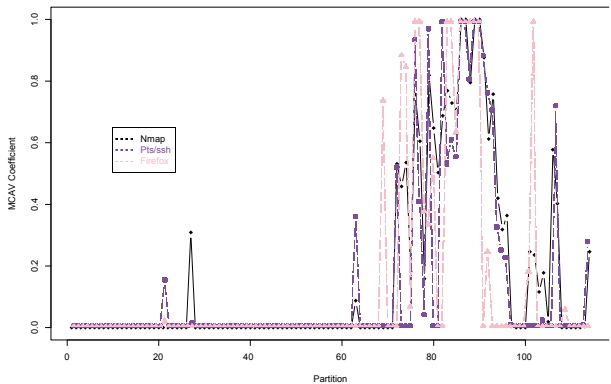

Figure 7: MCAV per 1000 antigens for the active normal dataset. Nmap, Pts and parent Firefox browser processes detected and represented.

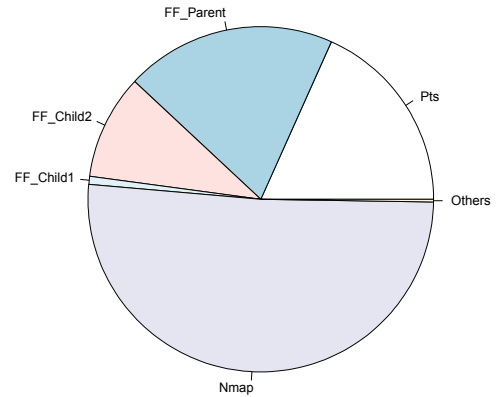

Figure 8: Proportionate chart of antigen per processes as input data for the active normal dataset

## 5. ANALYSIS

The results presented in Figure 6 show that the DCA can be used to detect large scale port-scans over an extended duration. It is important to note that the input data is both noisy and voluminous. The number of input antigen for both datasets is in excess of 1.3 million in total with the the actual input signal data also very noisy, with events such as network availability highly variable. Under 'night time' conditions, which are ideal, the algorithm performs remarkably well, compressing 1.3 million data items to a resultant 130,000 antigens (for PN), over the 7000s duration. The performance in terms of detection is also exemplary, indicating that the DCA is successful when applied to a 'real-world' scenario.

A number of false positives are shown through high MCAVs for the firefox process in the AN session, as shown in Figure 7. This indicates that the DCA has difficulty in separating concurrent normal and anomalous processes, when the analysis is based on the value of the MCAVs alone. However, the MCAV can be combined with the antigen confidence indicator. For example, in the AN experiment a total of 130,000 antigens are presented for analysis. Some 67000 of those antigens belong to the nmap process over the entire session, and can account for nearly 90% of the antigens produced when the nmap scan is highly active. In contrast, during periods of high nmap activity, the relative proportion of antigen presented belonging to the firefox process is under 10%. A combination of MCAV and antigen confidence indicatior can show not only how anomalous a process is deemed to be, but the level of confidence in the assessment. The greater the antigen input, the more times the antigen is sampled and the more accurate the MCAV. Charts representing the relative proportions of antigen input and antigen output are shown in Figures 8 and 9 respectively. Combining the relative proportion of antigen output per process with the MCAVs may lead to fewer false positives and effective anomaly detection.

Additionally, reduction of false positives may be achieved through exploring the facets of the number of antigens used to derive the MCAVs per process. In these experiments 10000 antigens were used for the calculation, but could be based on time or some other metric of system activity. As explored previously [4] the sampling DCs are sensitive to the length of time spent receiving signals in the tissue compartment. Longer sampling windows during low activity and shorter windows during high activity may make the detection more fine grained. Moving averages applied to the MCAV values over time may also alleviate this problem.

## 6. CONCLUSIONS AND FURTHER WORK

The DCA is a new development in AIS, and as yet has not been extensively tested. This paper presents work towards understanding the behaviour of the algorithm when applied to larger realistic problems. Its unique methods of combining multiple signals and correlating the combined values with a separate antigen data-stream works well for the detection of SYN scans over a long duration. However, some impairments in performance were shown when attempting to classify a scanning process when run concurrently with other active user-driven processes. Due to the nature of the input data and the methods of correlation employed, it is difficult to compare the algorithm with other standard methods, though in future network packet analysis may be performed for the sake of comparison. Additionally, individual signals alone are insufficient to produce successful classification based on both the volume of data and the amount of noise present in the input signals.

In addendum to the investigations proposed in Section 5, a number of future directions exist for the DCA. The first and most obvious future direction is a definitive benchmark test, to compare the performance of the DCA to other AIS and anomaly detection approaches. The introduction of adaptive signals or variable weights, for example using different weights at different times of the day, is another avenue to explore with the DCA. The algorithm may also be applied to other scan detection problems and to other problems in computer security. In addition applications outside of the scope of computer security can be considered such as the analysis of ra-dio data from space, or to mobile robotics. The results presented in this paper have shown that the DCA is capable of performing scan detection under difficult conditions through a unique form of immune inspired data fusion.

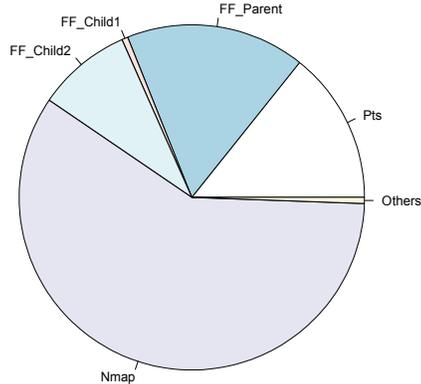

Figure 9: Proportionate chart of antigen per processes as output data for the active normal dataset

## 7. ACKNOWLEDGMENTS

This project is supported by the EPSRC (GR/S47809/01). Graphic design by Mark Hammonds.